\documentclass[conference]{IEEEtran}
\IEEEoverridecommandlockouts

\usepackage{cite}
\usepackage{amsmath,amssymb,amsfonts}
\usepackage{algorithmic}
\usepackage{graphicx}
\usepackage{textcomp}
\usepackage{xcolor}
\usepackage{fancyhdr}
\usepackage{hyperref}

\usepackage{float}
\usepackage{subcaption}
\usepackage{array}
\usepackage{booktabs}

\newcommand{\avr}{AVR}

\def\BibTeX{{\rm B\kern-.05em{\sc i\kern-.025em b}\kern-.08em
    T\kern-.1667em\lower.7ex\hbox{E}\kern-.125emX}}

\fancypagestyle{firstpage}{
  \fancyhf{}
  \fancyhead[C]{23rd Edition of IEEE India Council International Conference (INDICON 2026)}
  \fancyfoot[L]{%
      \begin{minipage}[t]{\textwidth}
      \normalsize{979-8-3195-2581-9/26/\$31.00 \copyright2026 IEEE}
    \end{minipage}%
  }
  
}

\begin{document}

\title{A Stage-Wise Explainable Pipeline for Arteriolar-to-Venular\\Ratio Estimation and Retinopathy Screening}


\author{
\IEEEauthorblockN{\mbox{%
Sharan Sharmeela Karuppannan$^{1\dagger}$,
Yoga Sri Varshan V$^{2\dagger}$,
Subin Sahayam$^{3}$,
Umarani Jayaraman$^{1*}$%
}}
\IEEEauthorblockA{
\begin{minipage}{0.98\textwidth}
\centering
\normalfont
$^{1}$Department of Computer Science and Engineering, Indian Institute of\\
Information Technology Design and Manufacturing, Kancheepuram,\\
Chennai, 600127, India.\\
$^{2}$The University of Texas at Austin, USA.\\
$^{3}$Department of Computer Science and Engineering, Shiv Nadar\\
University, Chennai, 603110, India.\\[1.5ex]
$^{*}$Corresponding author. E-mail:
\href{mailto:umarani@iiitdm.ac.in}{umarani@iiitdm.ac.in};\\
Contributing authors:
\href{mailto:sharansk18@gmail.com}{sharansk18@gmail.com};\\
\href{mailto:yogasrivan@utexas.edu}{yogasrivan@utexas.edu};
\href{mailto:subinsahayamm@snuchennai.edu.in}{subinsahayamm@snuchennai.edu.in};\\
$^{\dagger}$These authors contributed equally to this work.
\end{minipage}
}
}

\maketitle
\thispagestyle{firstpage}

\begin{abstract}
The arteriolar-to-venular diameter ratio (\avr) measured in the annulus surrounding the optic disc is a long-established clinical marker of retinal vascular health. Contemporary fundus screening systems largely bypass this marker and train end-to-end classifiers that map pixels directly to a diagnosis. Such classifiers provide limited visibility into the anatomical evidence underlying their predictions. We propose a five-stage pipeline that reconstructs the manual grading protocol used by ophthalmologists, with every stage emitting a human-readable artefact. The proposed pipeline combines optic disc localization, vessel calibre estimation using a minimal-chord boundary walk, and AVR computation from the dominant arterial and venular branches. Because every stage outputs an inspectable artefact, error can be attributed rather than merely observed. Evaluated on the HRF, LES-AV and MESSIDOR datasets, the pipeline attains 96.7\% accuracy and $F_1\!=\!0.968$ on HRF and 94.8\% accuracy and $F_1\!=\!0.922$ on MESSIDOR, without relying on a black-box end-to-end mapping. We argue that this attribution capability, and not just raw accuracy, is the reason to prefer procedural pipelines for clinical vascular biomarkers.
\end{abstract}


\section{Introduction}
\label{sec:intro}

The retina is the only site in the human body where an arterial and a venous bed can be photographed directly, non-invasively, and at low cost. This is particularly significant in resource-limited settings where diabetic retinopathy is highly prevalent but access to ophthalmologists is scarce. While fundus cameras are increasingly available in primary-health centres, the deep learning models used to interpret them are computationally intensive, typically requiring GPU-equipped servers or cloud infrastructure. Offloading sensitive medical images to remote servers raises data-privacy concerns, and GPU workstations are rarely available at the point of care. Our framework is designed to address both constraints, running entirely on a commodity CPU and processing a single fundus image (end-to-end) in under 4~seconds on a laptop, making it suitable for deployment on edge devices in resource-limited settings. Because retinal and cerebral microvasculature share embryological origin and barrier physiology, the calibre of retinal vessels carries information about vascular health beyond the eye. Narrowed retinal arterioles have been associated with hypertension, stroke, and diabetic complications \cite{wang2025geometry}.

The summary statistic most commonly used for this purpose is the arteriolar-to-venular ratio (AVR), defined as the calibre of the retinal arterioles divided by the calibre of the venules and measured within a standardised annulus around the optic disc \cite{keith1974some}. In healthy eyes, AVR is typically close to $2/3$ \cite{hubbard1999methods} and decreases as arteriolar narrowing develops or venular dilation occurs \cite{wang2025geometry}. Because AVR is derived directly from measurable vessel calibres, it provides a mechanistically interpretable biomarker of retinal vascular health.
\vspace{-0.85mm}
\subsection{The case against end-to-end classification}

Almost all recent automated fundus screening work discards this structure.
A convolutional or transformer backbone is trained on labelled fundus images
and outputs a diagnosis directly.
However, these models remain difficult to deploy reliably in clinical settings due to the following reasons:

\textbf{Training bias and shortcut learning.} A network optimised only for the label has no reason to attend to vessel calibre if a cheaper cue correlates with the label in the training set. Networks routinely exploit dataset-specific cues such as scanner signatures and image processing chains instead of pathology \cite{geirhos2020shortcut}. Because such shortcuts are dataset-specific, performance collapses when the acquisition
device, population or illumination changes. Similar effects have been observed under cross-dataset evaluation of retinal vessels \cite{sharan2025avseg}.

\textbf{Absence of an inspectable intermediate.} An end-to-end classifier
that outputs ``retinopathy, $p=0.87$'' offers a clinician nothing to
verify. Saliency maps are post-hoc and are not faithful to the computation
\cite{rudin2019stop}. If the prediction is wrong there is no stage to blame,
no artefact to correct, and no way to distinguish a segmentation failure
from an ambiguous eye.

\subsection{What we propose}

We propose a systematic, procedural alternative that computes \avr\ the way a
trained grader does, and derives the diagnosis from that number alone. The
pipeline has five stages:

\begin{enumerate} \item \textbf{Input.} A colour fundus photograph. \item \textbf{Dual-mask segmentation.} Vessel and A/V masks. \item \textbf{Optic disc estimation.} \item \textbf{Vessel calibre measurement.} Minimal-chord boundary-walk calibre estimation \cite{varshan2023diameter}. \item \textbf{Ratio and decision.} \avr\ computation and threshold-based classification. \end{enumerate}
Unlike conventional end-to-end CNN and transformer-based screening systems, the proposed pipeline preserves the clinical measurement process from image acquisition to AVR estimation. The final decision is derived from explicit anatomical measurements rather than image features, allowing predictions to be traced through a transparent decision pathway.
\subsection{Why this matters more than the headline accuracy}

The practical payoff of a transparent pipeline is that its errors are attributable. Any abstention or misclassification can be traced to a specific geometric failure, such as optic disc localization, artery--vein assignment, or calibre measurement.

\subsection{Contributions}

\begin{itemize}
  \item A five-stage, training-free-at-the-measurement-level pipeline for
        \avr\ estimation that reproduces the grading protocol, with
        an inspectable artefact at every stage.
  \item A decoupled dual-mask design that separates boundary estimation from
        A/V classification, with a label-transfer rule that
        resolves disagreement between the two masks.
  \item Demonstration on LES-AV, HRF and MESSIDOR with precise cross-sectional analysis of the proposed pipeline and its methodology.
  \item We propose the first end-to-end diagnostic framework
        for retinal disease screening that is both modular and
        inspectable at every stage. Each component is independently replaceable, enabling extension to conditions
        beyond retinopathy, to other optic-nerve and vascular pathologies,
        without retraining the system.
\end{itemize}

\section{Related Work}
\label{sec:related}

\subsection{End-to-end retinopathy classification}The dominant paradigm maps fundus images directly to diagnostic labels using convolutional or transformer architectures. However, these systems often fail when deployment distributions shift due to shortcut learning \cite{geirhos2020shortcut}, where models key on dataset-specific cues rather than pathology. Rudin \cite{rudin2019stop} argues that high-stakes domains require inherently interpretable models over post-hoc explanations.

\subsection{Artery--vein segmentation}Reliable AVR estimation requires precise artery and vein identification. While early methods were graph-theoretic, modern approaches rely on deep learning \cite{hemelings2019artery}. We adopt an Attention-WNet architecture \cite{sharan2025avseg} that trains separate binary models for arteries and veins rather than a single multi-class model. This separate-model design is essential to the dual-mask framework (Section~\ref{sec:dualmask}) used to decouple geometry from class assignment.

\subsection{Automated vessel calibre and \avr\ measurement}A parallel literature focuses on measuring vessel calibre and automating AVR computation from fundus photographs. We employ the minimal-chord boundary walk \cite{varshan2023diameter}, which estimates calibre by fixing one boundary point and traversing the Moore neighbourhood to find the locally minimal chord. Recent systems such as VascX \cite{vargas2025vascx} scale retinal vascular analysis and calibre quantification for large datasets.

\subsection{Position of this work}Our approach sits at the intersection of these domains: it consumes learned artery--vein masks, applies a classical calibre algorithm, and outputs an interpretable clinical decision. Unlike profiling tools such as VascX \cite{vargas2025vascx}, we treat the pipeline as an inspectable clinical tool. Furthermore, we use it as an object of study to quantify exactly how explicitly modeled geometry drives screening classifications.

\section{Clinical Basis and Problem Formulation}
\label{sec:clinical}

\subsection{The measurement annulus}
Retinal vessels taper continuously as they radiate from the optic disc, so a calibre measurement is meaningless without a specified location. The convention established by the Atherosclerosis Risk in Communities protocol \cite{hubbard1999methods} partitions the peripapillary retina into concentric annuli indexed by distance from the optic disc margin. Zone~B (0.5--1.0 disc diameters from the margin) is the standard measurement zone, and all normative AVR data and published diagnostic thresholds are calibrated to it. Equivalently, with $r_d$ the disc radius and the disc centre as origin, Zone~B is the annulus with inner radius $2r_d$ and outer radius $3r_d$. The choice of zone is not cosmetic. Wang et al.\ \cite{wang2025geometry} found AVR in Zone~B significantly lower in silent brain infarction cases than in matched controls ($0.72\pm0.06$ versus $0.73\pm0.08$, $p=0.023$), while the same ratio measured in Zone~C showed no significant difference ($p=0.182$). We therefore restrict all measurements to Zone~B.

\subsection{Definition of \avr} Let $\mathcal{A}=\{a_1,a_2,\dots\}$ and $\mathcal{V}=\{v_1,v_2,\dots\}$ denote the arterial and venular calibres crossing Zone~B, sorted in decreasing order. Classical AVR computation uses CRAE and CRVE derived from the six largest vessels of each type using the revised Knudtson formula \cite{knudtson2003revised}. In fully automatic artery--vein masks, however, six confidently labelled vessels are often unavailable. We therefore form two dominant pairings and compute a per-pair ratio:
\begin{equation}
  \label{eq:avr}
  \avr_i = \frac{a_i}{v_i}, \qquad i = 1,2
\end{equation}
where $(a_1,v_1)$ and $(a_2,v_2)$ are the largest and second-largest artery--vein pairs, respectively. This focuses on the most reliably segmented vessels while preserving traceability to each individual measurement.

\subsection{Decision rule}

A healthy \avr\ is close to $2/3$
\cite{keith1974some,hubbard1999methods}. Since asymmetric arteriolar narrowing can affect a single vessel pairing while
the other remains normal, we flag an image as abnormal whenever the
\emph{worse} pairing falls below the clinical threshold:

\begin{equation}
  \label{eq:decision}
  \hat{y} =
  \begin{cases}
    \text{normal} & \text{if } \min(\avr_1,\,\avr_2) \geq \tau, \\
    \text{retinopathy} & \text{if } \min(\avr_1,\,\avr_2) < \tau,
  \end{cases}
  \qquad \tau = 0.67 .
\end{equation}

The threshold $\tau=0.67$ was prespecified before evaluation, adopted from the clinical convention that a healthy \avr\ is approximately $2/3$ \cite{keith1974some,hubbard1999methods}. We note that conventional \avr\ computed from six vessels and our two-pair estimator are different quantities, so $\tau=0.67$ should be regarded as a prespecified operating point rather than a clinically calibrated cutoff. Table~\ref{tab:threshold_sweep} shows that this value also attains the highest Youden index among evaluated thresholds on the HRF cohort.

\subsection{Abstention}

If Stage~4 recovers fewer than two arteries or fewer than two veins in
Zone~B, Eq.~\eqref{eq:avr} is undefined and the pipeline abstains rather than
emitting a value. Abstention is a feature of a procedural system: the reason
for abstention is always a concrete, visible deficiency in an earlier
artefact. We report coverage alongside every accuracy number.

\section{Method}
\label{sec:method}

\subsection{Overview}

\begin{figure*}[t]
  \centering
  \includegraphics[width=0.7\textwidth]{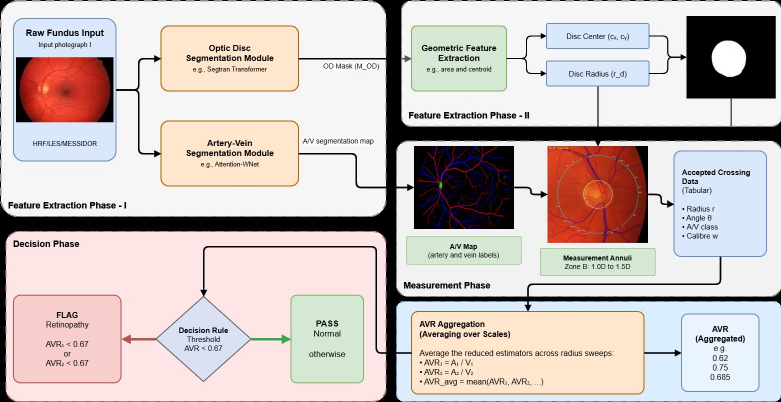}
  \vspace{1mm}

  \begin{tabular}{ccccc}
    \includegraphics[width=0.13\textwidth]{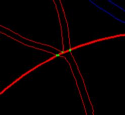} &
    \includegraphics[width=0.13\textwidth]{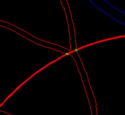} &
    \includegraphics[width=0.13\textwidth]{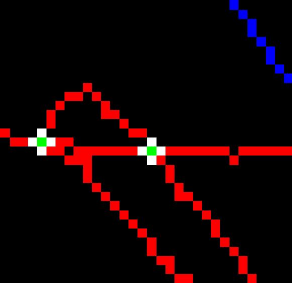} &
    \includegraphics[width=0.13\textwidth]{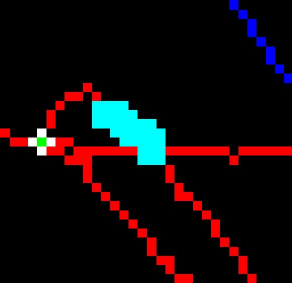} &
    \includegraphics[width=0.13\textwidth]{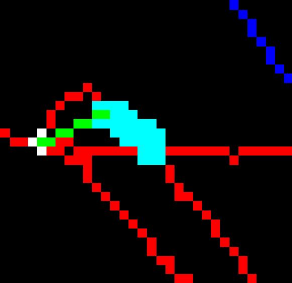} \\
    (a) Branch & (b) Radius & (c) Initial & (d) Walk & (e) Chord
  \end{tabular}
  \caption{The five-stage pipeline and Stage~4 calibre measurement.
    (a,b)~Branch handling: an odd crossing count indicates bifurcation;
    the radius is reduced by $0.01r_d$ until parity is restored.
    (c--e)~Minimal-chord boundary walk: B migrates along the vessel
    boundary until AB is shortest, converting an oblique chord into
    vessel width.}
  \label{fig:pipeline}
\end{figure*}

Figure~\ref{fig:pipeline} shows the pipeline. An input photograph $I$ passes
through five stages, each of which writes an artefact to disk:
a preprocessed image, two segmentation masks, a disc circle, a table of
measured chords, and a scalar ratio with a decision. The artefacts are the
explanation; there is no separate explanation module.

\subsection{Stage 1: Input and preprocessing}
\label{sec:preproc}

The pipeline accepts a colour fundus photograph, preferably disc-centred. The
field of view is isolated by thresholding the image intensity and retaining
the largest connected component, which removes the black border and prevents
it from being mistaken for vascular structure later.

Only the green channel is retained for segmentation input, as it provides the highest contrast between haemoglobin-bearing vessels and the retinal background and the sharpest separation between arterial and venous reflectance \cite{sharan2025avseg}. Contrast-limited adaptive histogram equalisation (tile grid $8\times8$, clip limit $2.0$) is applied, followed by per-image intensity standardisation to zero mean and unit variance. HRF images ($3504\times2336$) are resized to $1024\times1024$ before segmentation to reduce CPU cost. Because all geometric quantities (Zone~B radii, vessel widths) are defined relative to the optic disc radius $r_d$, the resulting \avr\ is a dimensionless ratio that is invariant to input resolution.

\subsection{Stage 2: Decoupled dual-mask segmentation}
\label{sec:dualmask}

Stage~2 produces two masks.

\textbf{The binary vessel mask $M_{v}$} is produced by a model trained only to
separate vessel from background. Its role is purely geometric: it defines
where the vessel boundary lies, and therefore every calibre the pipeline
will report.

\textbf{The artery--vein mask $M_{av}$} is produced by the Attention-WNet of Sharan et al.\ \cite{sharan2025avseg}. Two binary Attention-WNet models are used, one for arteries and one for veins; each pixel is assigned to the stronger response and marked \emph{uncertain} when the activations differ by less than $20\%$. Its role is semantic: it determines vessel type, not vessel boundary.

\subsubsection{Label transfer}
The two masks are produced independently and may disagree at individual pixels. Rather than intersecting the masks pixelwise, labels from $M_{av}$ are transferred to connected components of $M_v$.

Let $C$ be a connected component of $M_{v}$ that intersects the Zone~B
annulus. Let $n_{a}(C)$, $n_{v}(C)$ and $n_{u}(C)$ be the number of pixels of
$C$ labelled artery, vein and uncertain by $M_{av}$. The component label is assigned using
Eq.~\eqref{eq:labeltransfer}:

\begin{equation}
  \label{eq:labeltransfer}
  \ell(C) =
  \begin{cases}
    \text{artery} & \text{if } n_{a} > n_{v} \text{ and } n_{a} \geq \rho \, (n_{a} + n_{v}), \\
    \text{vein}   & \text{if } n_{v} > n_{a} \text{ and } n_{v} \geq \rho \, (n_{a} + n_{v}), \\
    \text{unassigned} & \text{otherwise,}
  \end{cases}
\end{equation}

with agreement threshold $\rho = 0.6$. Voting over a whole component, rather than per pixel, exploits a fact that
pixelwise intersection throws away: a vessel segment does not change type
along its length. It also makes the assignment robust to the crossing points
where $M_{av}$ is least reliable, since those contribute a minority of the
component's pixels.

\subsection{Stage 3: Optic disc estimation} \label{sec:od} Stage~3 estimates the disc centre $c=(c_x,c_y)$ and radius $r_d$, which define the coordinate frame and scale for all subsequent measurements. An intensity-based detector identifies the brightest large structure in the fundus image through red-channel filtering and thresholding, while a Segtran-based detector directly predicts a disc mask from which $c$ and $r_d$ are obtained. Given $c$ and $r_d$, Zone~B is defined as $r \in [2r_d,3r_d]$. Vessel measurements are sampled across this interval rather than on a single measurement circle.

\subsection{Stage 4: Vessel calibre measurement}
\label{sec:calibre}

Stage~4 measures the width of every vessel crossing a measurement circle using  the algorithm of Yoga Sri Varshan et al.\ \cite{varshan2023diameter}, which has been adapted as follows, for the dual-mask representation and a radius sweep.

\subsubsection{Boundary extraction and intersection}

A Laplacian operator is applied to $M_{v}$ to extract the vessel boundary
$\partial M_{v}$. A measurement circle of radius
$r$ centred at $c$ is rasterised and intersected with $\partial M_{v}$. Rasterising the circle at one-pixel width misses intersections whenever the discretised circle and boundary run parallel without sharing a pixel. We therefore rasterise at two-pixel width, which returns a small \emph{cluster} of boundary pixels at each crossing. Since every cluster is a connected component of the intersection set, one representative per component recovers one point per crossing.

\subsubsection{Pairing and branch handling}

The recovered points are sorted by the angle they subtend at $c$ and paired
consecutively, so that each pair spans one vessel. A branch point on the measurement circle produces an odd number of crossings, causing consecutive pairing to straddle different vessels and report a spurious width. The parity of the crossing count is therefore a sufficient detector. If the count is odd, the radius is reduced by $0.01r_d$ and the intersection recomputed until parity is restored. Because branch points are isolated, a small perturbation moves the circle off the bifurcation (Fig.~\ref{fig:pipeline}).

\subsubsection{Minimal-chord boundary walk}

Let $A$ and $B$ be a paired set of boundary points. The Euclidean distance
$\lVert A - B \rVert$ overestimates the vessel width whenever the circle crosses the vessel obliquely, and the error grows with the obliquity
angle. The correction holds $A$ fixed and walks $B$ along the vessel boundary
to the point that minimises the chord. At each iteration, the eight Moore neighbours of $B$ that lie on
$\partial M_{v}$ are examined and $B$ moves to whichever minimises
$\lVert A - \cdot \rVert$. The walk terminates when no neighbour improves on
the incumbent, i.e.\ when $B$ is a local minimiser of distance to $A$ along
the boundary. The Moore neighbourhood is used in preference to the
von Neumann neighbourhood because an 8-connected boundary walk can follow
diagonal vessel edges, which a 4-connected walk cannot (Fig.~\ref{fig:pipeline}).

\subsubsection{Quality filters}

A measured chord is discarded if it falls outside a plausible calibre range, if its endpoints belong to different connected components of $M_v$, or if the supporting vessel segment is shorter than a minimum length. Every rejection is logged with its reason, allowing the pipeline to abstain rather than report an invalid measurement.

\subsection{Stage 5: Ratio, aggregation and decision}
\label{sec:aggregation}

\subsubsection{Radius aggregation}

A single measurement circle may coincide with a branch point or
crossing. To mitigate this, we sweep $r$ across $[2r_d,3r_d]$ in
steps of $0.05r_d$ and compute Eq.~\eqref{eq:avr} at each radius,
discarding any radius that yields fewer than two valid pairs. The
final $\avr_i$ for each pairing $i$ is the median of $\avr_i(r)$
over the surviving radii, and the screening decision applies
Eq.~\eqref{eq:decision} to $\min(\avr_1, \avr_2)$.

\subsubsection{Decision and audit trail}

The aggregated \avr\ is thresholded at $\tau=0.67$ per Eq.~\eqref{eq:decision}. Alongside the decision, the pipeline emits all intermediate artefacts, including segmentation masks, optic-disc estimates, calibre measurements, rejection logs, and per-radius \avr\ values.

\section{Experimental Setup}
Table~\ref{tab:setup} summarises the datasets, evaluation metrics, training configuration, and fixed geometric parameters used throughout the experiments. Dataset resolutions are reported to support cross-dataset evaluation across substantially
different scales.

\begin{table}[h] \centering \small \setlength{\tabcolsep}{3pt} \caption{Experimental setup.} \label{tab:setup} \begin{tabular}{p{1.9cm}p{5.0cm}} \toprule \textbf{Item} & \textbf{Description} \\ \midrule Datasets & DRIVE \cite{staal2004ridge} (40, $565\times584$) for A/V training; HRF \cite{budai2013robust} (45, $3504\times2336$, 30 evaluated) for screening; LES-AV \cite{orlando2018towards} (22, $1620\times1444$) for calibre validation; MESSIDOR \cite{decenciere2014feedback} (1200, $\leq2240\times1488$, 800 evaluated) for large-scale analysis. \\ Screening Metrics & Accuracy, Precision, Recall, $F_1$, AUC. \\ Segmentation Metrics & All-pixel, centreline, and wide-vessel centreline accuracy/$F_1$ \cite{hemelings2019artery}. \\ Training & Adam ($lr=0.01$), batch size 6, 200 epochs, patience 20, focal loss ($\gamma=2$, $\alpha=0.8/0.9$) \cite{sharan2025avseg}. \\ Implementation & NumPy/SciPy geometry; CPU only. \\ Fixed Parameters & $\rho=0.6$, radius step $0.05r_d$, branch step $0.01r_d$, quality-filter bounds fixed before evaluation. \\ \bottomrule \end{tabular} \end{table}

\section{Results}
\label{sec:results}

\subsection{Stage 2: Artery--vein segmentation}
\label{sec:res-seg}
Table~\ref{tab:seg} reproduces the performance of the Attention-WNet artery--vein stage \cite{sharan2025avseg}. Two observations matter for the downstream pipeline. First, the gap between all-pixel accuracy ($0.960$ on HRF) and wide-vessel centreline accuracy ($0.845$) confirms the asymmetry that motivated the dual-mask design in Section~\ref{sec:dualmask}: the model is far better at saying \emph{that} a pixel is vascular than at resolving class along the vessel. Second, cross-dataset transfer is comparatively graceful --- $0.939$ all-pixel accuracy in both transfer directions, with per-class artery and vein accuracies above $0.959$ --- which is the property we require, since the downstream calibre stage is applied to clinical images the segmentation model has not seen.

\begin{table}[h]
  \centering
  \caption{Artery--vein segmentation accuracy and $F_1$ for the Stage~2 model \cite{sharan2025avseg}. ``Wide centreline'' denotes centreline pixels of vessels wider than two pixels, the regime the calibre stage operates in.}
  \label{tab:seg}
  \small
  \begin{tabular}{lcccc}
    \toprule
     & \multicolumn{2}{c}{All vessel pixels} & \multicolumn{2}{c}{Wide centreline} \\
    \cmidrule(lr){2-3} \cmidrule(lr){4-5}
    Setting & Accuracy & $F_1$ & Accuracy & $F_1$ \\
    \midrule
    HRF (within) & $0.960$ & $0.958$ & $0.845$ & $0.845$ \\
    DRIVE (within) & $0.950$ & $0.946$ & $0.780$ & $0.778$ \\
    HRF $\rightarrow$ DRIVE & $0.939$ & $0.925$ & $0.736$ & $0.724$ \\
    DRIVE $\rightarrow$ HRF & $0.939$ & $0.940$ & $0.712$ & $0.707$ \\
    \bottomrule
  \end{tabular}
\end{table}

\subsection{Baseline Calibre Validation on LES-AV}
\label{sec:res-les}
To evaluate the geometric stage using reference masks, we ran the full sweep pipeline on all 22 LES-AV images using the A/V masks provided with the dataset. All 22 images produced valid measurements (100\% coverage). The median \avr\ across images was 0.818 (95\% CI: [0.792, 0.978]), and 20 of 22 images yielded min($\avr_1$, $\avr_2$) $\geq$ 0.67. Since LES-AV contains 11 healthy and 11 glaucomatous subjects rather than retinopathy-labelled cohorts, these results demonstrate measurement availability and characterise the resulting \avr\ distribution, but do not constitute retinopathy evaluation.

\subsection{Retinopathy Screening on HRF}
\label{sec:res-hrf}

\begin{figure*}[t]
  \centering
  \begin{subfigure}[c]{0.22\textwidth}
    \centering
    \includegraphics[width=0.7\linewidth]{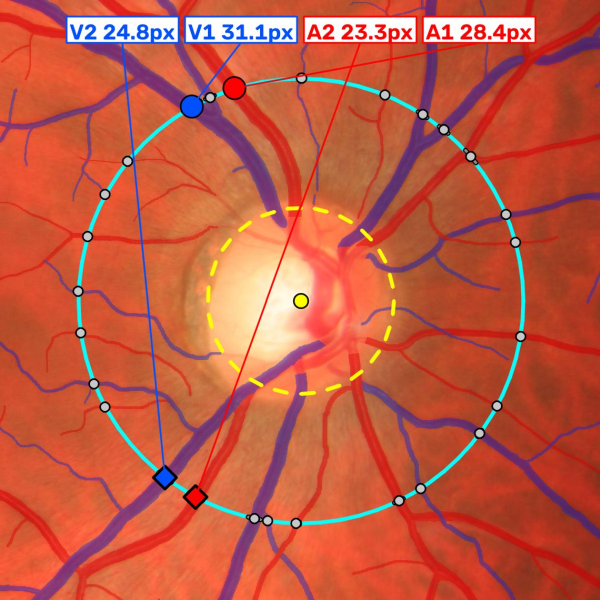}
    \caption{\texttt{01\_h}}
  \end{subfigure}\hspace{0.01\textwidth}
  \begin{subfigure}[c]{0.22\textwidth}
    \centering
    \includegraphics[width=0.7\linewidth]{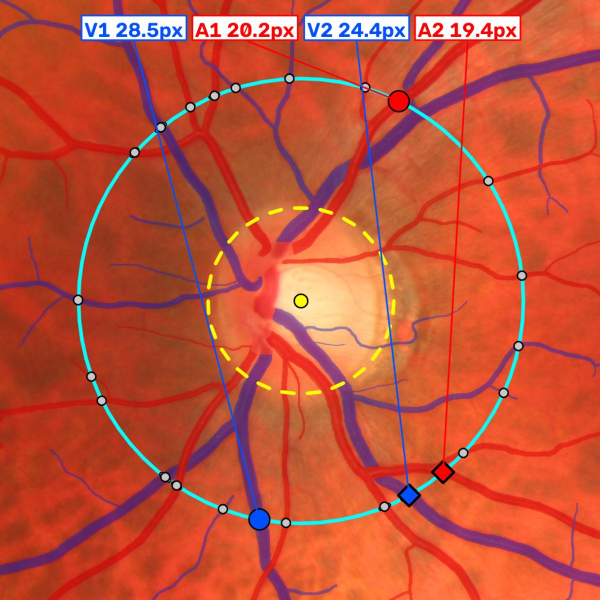}
    \caption{\texttt{02\_h}}
  \end{subfigure}\hspace{0.01\textwidth}
  \begin{subfigure}[c]{0.22\textwidth}
    \centering
    \includegraphics[width=0.7\linewidth]{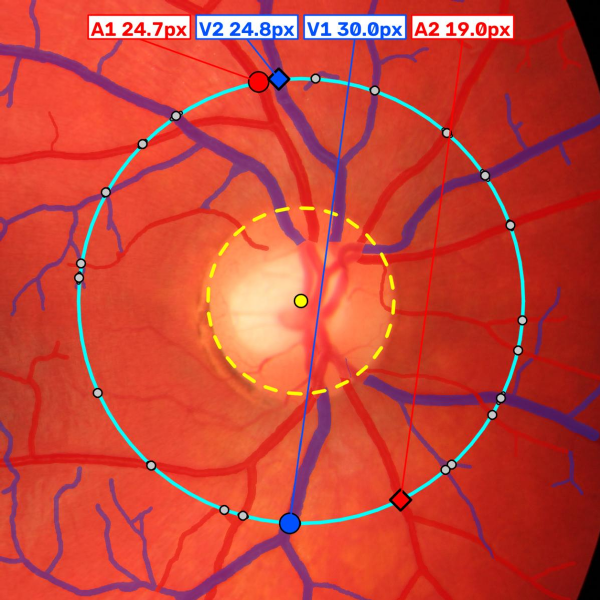}
    \caption{\texttt{03\_h}}
  \end{subfigure}\hspace{0.01\textwidth}
  \begin{minipage}[c]{0.28\textwidth}
    \centering
    \scriptsize
    \textbf{Healthy ($\avr > 0.67$)}\\[2pt]
    \begin{tabular}{lccc}
      \toprule
      Case & $\avr_1$ & $\avr_2$ & Mean \\
      \midrule
      01\_h & 0.911 & 0.939 & \textbf{0.925} \\
      02\_h & 0.707 & 0.796 & \textbf{0.752} \\
      03\_h & 0.825 & 0.768 & \textbf{0.796} \\
      \bottomrule
    \end{tabular}
  \end{minipage}

  \vspace{0.3em}

  \begin{subfigure}[c]{0.22\textwidth}
    \centering
    \includegraphics[width=0.7\linewidth]{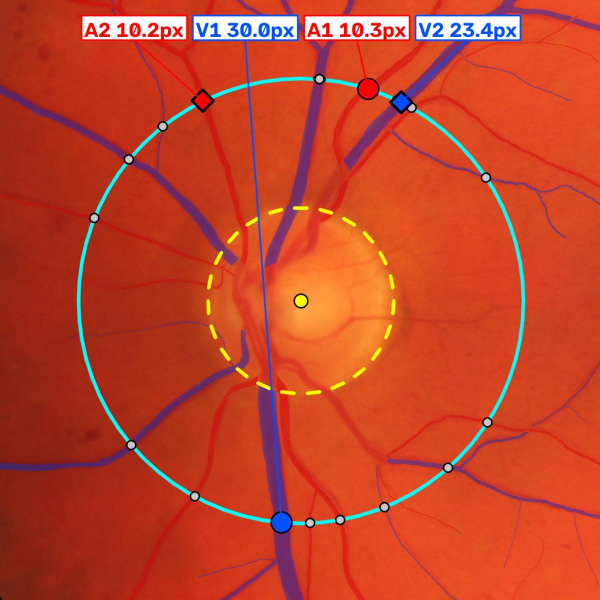}
    \caption{\texttt{01\_dr}}
  \end{subfigure}\hspace{0.01\textwidth}
  \begin{subfigure}[c]{0.22\textwidth}
    \centering
    \includegraphics[width=0.7\linewidth]{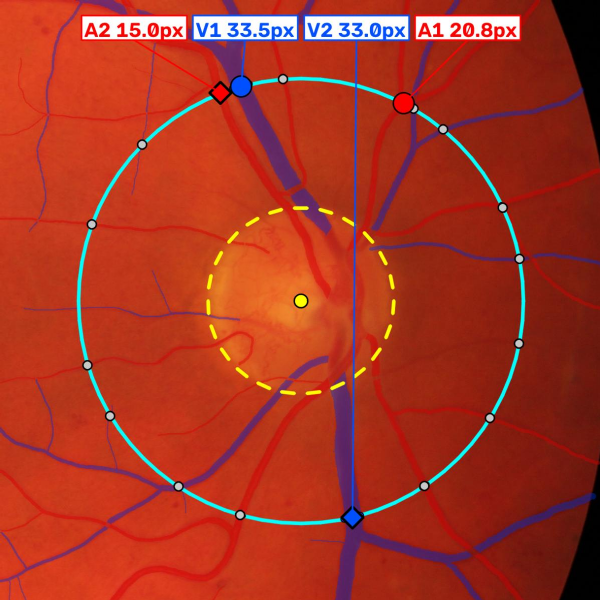}
    \caption{\texttt{02\_dr}}
  \end{subfigure}\hspace{0.01\textwidth}
  \begin{subfigure}[c]{0.22\textwidth}
    \centering
    \includegraphics[width=0.7\linewidth]{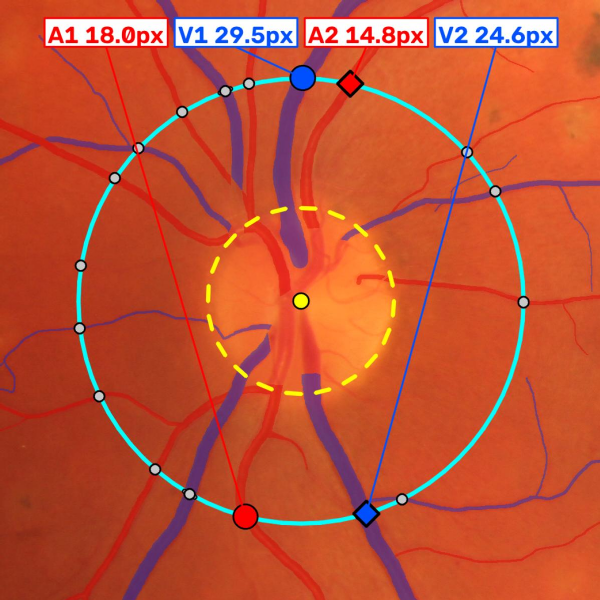}
    \caption{\texttt{03\_dr}}
  \end{subfigure}\hspace{0.01\textwidth}
  \begin{minipage}[c]{0.28\textwidth}
    \centering
    \scriptsize
    \textbf{Retinopathy ($\avr < 0.67$)}\\[2pt]
    \begin{tabular}{lccc}
      \toprule
      Case & $\avr_1$ & $\avr_2$ & Mean \\
      \midrule
      01\_dr & 0.342 & 0.437 & \textbf{0.390} \\
      02\_dr & 0.621 & 0.454 & \textbf{0.538} \\
      03\_dr & 0.611 & 0.603 & \textbf{0.607} \\
      \bottomrule
    \end{tabular}
  \end{minipage}

  \caption{\textbf{Representative healthy and retinopathy cases with corresponding AVR measurements. In the examples shown, the minimum of AVR$_1$ and AVR$_2$ is $\geq$ 0.67 for healthy cases and $< $ 0.67 for retinopathy cases. The arithmetic mean is shown for descriptive purposes only; classification uses the minimum of the two ratios.}}
  \label{fig:hrf_6_images_detailed}
\end{figure*}

We evaluated the end-to-end pipeline on the 30-image HRF screening cohort (15 healthy and 15 diabetic retinopathy eyes). A fundus image is classified as retinopathic if either of the dominant vessel pairings ($\avr_1$ or $\avr_2$) falls below the clinical threshold of 0.67 (Eq.~\eqref{eq:decision}). The extracted \avr\ measurements clearly separated the two cohorts, with one false positive and no false negatives (Table~\ref{tab:hrf_unified}). The area under the ROC curve is $\text{AUC}=0.978$, and the Youden-optimal threshold coincides with the clinical value $\tau=0.67$ ($J=0.933$). Table~\ref{tab:threshold_sweep} reports the full threshold analysis. Fig.~\ref{fig:hrf_6_images_detailed} shows representative
healthy and diabetic retinopathy cases with their AVR measurements.

\begin{table}[h]
  \centering
  \caption{Comprehensive HRF evaluation summary ($N=30, \tau=0.67$). This unified view presents the geometric class separation, the resulting confusion matrix, and the overall diagnostic metrics. 95\% confidence intervals (Wilson for proportions, bootstrap for $F_1$ and AUC) are reported in brackets.}
  \label{tab:hrf_unified}
  \resizebox{\columnwidth}{!}{%
  \begin{tabular}{lcc|cc}
    \toprule
    & & & \multicolumn{2}{c}{\textbf{Predicted Class}} \\
    \textbf{Actual Cohort} & \textbf{Images} & \textbf{Mean \avr} & \textbf{Healthy} & \textbf{Retinopathy} \\
    \midrule
    Healthy & 15 & $0.789$ & \textbf{14} & 1 \\
    Diabetic Retinopathy & 15 & $0.624$ & 0 & \textbf{15} \\
    \midrule
    \multicolumn{3}{l}{\small \textbf{Accuracy:} 96.67\% [83.3, 99.4]} & \multicolumn{2}{l}{\small \textbf{Sensitivity:} 100.0\% [79.6, 100]} \\
    \multicolumn{3}{l}{\small \textbf{$F_1$-Score:} 0.968 [0.889, 1.0]} & \multicolumn{2}{l}{\small \textbf{Specificity:} 93.33\% [70.2, 98.8]} \\
    \multicolumn{3}{l}{\small \textbf{AUC:} 0.978 [0.919, 1.0]} & \multicolumn{2}{l}{\small \textbf{Precision:} 93.75\% [71.7, 98.9]} \\
    \bottomrule
  \end{tabular}%
  }
\end{table}

\begin{table}[h]
  \centering
\caption{Threshold sensitivity analysis on HRF ($N=30$). $J$ denotes Youden's index ($J = \text{Sens} + \text{Spec} - 1$). The Youden-optimal threshold coincides with the clinical value $\tau=0.67$.}
  \label{tab:threshold_sweep}
  \small
  \begin{tabular}{ccccccc}
    \toprule
    $\tau$ & Acc & Sens & Spec & Prec & $F_1$ & $J$ \\
    \midrule
    0.50 & 0.567 & 0.133 & 1.000 & 1.000 & 0.235 & 0.133 \\
    0.55 & 0.600 & 0.200 & 1.000 & 1.000 & 0.333 & 0.200 \\
    0.60 & 0.733 & 0.467 & 1.000 & 1.000 & 0.636 & 0.467 \\
    0.65 & 0.833 & 0.733 & 0.933 & 0.917 & 0.815 & 0.667 \\
    \textbf{0.67} & \textbf{0.967} & \textbf{1.000} & \textbf{0.933} & \textbf{0.938} & \textbf{0.968} & \textbf{0.933} \\
    0.70 & 0.900 & 1.000 & 0.800 & 0.833 & 0.909 & 0.800 \\
    0.75 & 0.800 & 1.000 & 0.600 & 0.714 & 0.833 & 0.600 \\
    0.80 & 0.767 & 1.000 & 0.533 & 0.682 & 0.811 & 0.533 \\
    0.85 & 0.567 & 1.000 & 0.133 & 0.536 & 0.698 & 0.133 \\
    \bottomrule
  \end{tabular}
\end{table}

\subsection{Ablation Study}
\label{sec:ablation}

To assess the contribution of each design choice, we ablate four axes
on the 30-image HRF cohort at $\tau=0.67$ using the
$\min(\avr_1,\avr_2)$ rule (Table~\ref{tab:ablation}).

\textbf{Single radius vs.\ sweep.}
A single measurement circle at $r\!=\!2.25R_d$ achieves 95.2\% accuracy
but only 21/30 coverage; other radii fare worse. The sweep over
$[2R_d,3R_d]$ recovers all 30 images and reaches 96.7\% accuracy,
confirming that median aggregation across Zone~B is essential for
robustness.

\textbf{Top-$k$ vessel pairs.}
Using only the single largest pairing ($k\!=\!1$) drops accuracy to
86.7\% (three DR cases missed). $k\!\in\!\{2,3,4\}$ all yield identical
96.7\% accuracy; $k\!=\!6$ degrades to 90.0\% because few images have
six reliably labelled arteries. We adopt $k\!=\!2$ as the most
conservative choice.

\textbf{Euclidean vs.\ minimal-chord.}
Replacing the boundary walk with Euclidean distance reduces accuracy to 90.0\% and sensitivity to 86.7\%, confirming
that the obliquity correction of the minimal-chord walk matters for
screening.

\textbf{Label transfer vs.\ pixelwise pairing.}
Replacing vessel-aware label transfer with angular pairing reduces specificity to 80.0\%, as angular pairing matches intersection points by adjacency and can mispair points on different vessels.

\begin{table}[h]
  \centering
  \caption{Ablation study on HRF ($N\!=\!30$, $\tau\!=\!0.67$,
  $\min(\avr_1,\avr_2)$ rule). Bold marks the chosen configuration.}
  \label{tab:ablation}
  \small
  \resizebox{0.9\columnwidth}{!}{%
  \begin{tabular}{lccccc}
    \toprule
    Configuration & $N_\text{valid}$ & Acc & Sens & Spec & $F_1$ \\
    \midrule
    \multicolumn{6}{l}{\textit{Radius strategy (top-$k\!=\!2$, chord)}} \\
    \quad Single $r\!=\!2.00R_d$ & 23 & 0.783 & 0.750 & 0.818 & 0.783 \\
    \quad Single $r\!=\!2.25R_d$ & 21 & 0.952 & 1.000 & 0.900 & 0.957 \\
    \quad Single $r\!=\!2.50R_d$ & 20 & 0.850 & 0.900 & 0.800 & 0.857 \\
    \quad \textbf{Sweep $[2R_d,3R_d]$} & \textbf{30} & \textbf{0.967} & \textbf{1.000} & \textbf{0.933} & \textbf{0.968} \\
    \midrule
    \multicolumn{6}{l}{\textit{Top-$k$ pairs (sweep, chord)}} \\
    \quad $k=1$ & 30 & 0.867 & 0.800 & 0.933 & 0.857 \\
    \quad \textbf{$k=2$} & \textbf{30} & \textbf{0.967} & \textbf{1.000} & \textbf{0.933} & \textbf{0.968} \\
    \quad $k=3$ & 30 & 0.967 & 1.000 & 0.933 & 0.968 \\
    \quad $k=4$ & 30 & 0.967 & 1.000 & 0.933 & 0.968 \\
    \quad $k=6$ & 30 & 0.900 & 0.867 & 0.933 & 0.897 \\
    \midrule
    \multicolumn{6}{l}{\textit{Width estimator (sweep, $k\!=\!2$)}} \\
    \quad Euclidean (no walk) & 30 & 0.900 & 0.867 & 0.933 & 0.897 \\
    \quad \textbf{Minimal-chord (walk)} & \textbf{30} & \textbf{0.967} & \textbf{1.000} & \textbf{0.933} & \textbf{0.968} \\
    \midrule
    \multicolumn{6}{l}{\textit{Pairing method (sweep, $k\!=\!2$, chord)}} \\
    \quad Angular (pixelwise) & 30 & 0.900 & 1.000 & 0.800 & 0.909 \\
    \quad \textbf{Vessel-aware (label transfer)} & \textbf{30} & \textbf{0.967} & \textbf{1.000} & \textbf{0.933} & \textbf{0.968} \\
    \bottomrule
  \end{tabular}%
  }
\end{table}

\subsection{Exploratory Comparison with VascX}
\label{sec:res-vascx}

To provide a quantitative reference point, we evaluated VascX \cite{vargas2025vascx} (v2.1.0) on the same 30-image HRF cohort. VascX computes Zone~B CRAE and CRVE from up to six vessels using the Knudtson recursion; we applied the same prespecified $\tau=0.67$ to the resulting CRAE/CRVE ratio without threshold fitting. A post-hoc threshold sweep over VascX's outputs identified $\tau=0.62$ as the Youden-optimal operating point for its estimator. Both systems achieved 100\% coverage (30/30). Table~\ref{tab:vascx} compares the two systems.

At the shared $\tau\!=\!0.67$, the proposed pipeline produces 1 false positive versus 8 for VascX, and a paired accuracy advantage of 23.3 percentage points (bootstrap 95\% CI: [10.0, 36.7]). VascX does not reach the proposed pipeline's 96.7\% even at its own Youden-optimal threshold.

Three caveats limit interpretation: (i)~VascX development data includes HRF, whereas the proposed pipeline's segmentation models were trained exclusively on DRIVE, so HRF is unseen data only for our system; (ii)~VascX computes a six-vessel Knudtson AVR, a different quantity from our two-pair minimum estimator; (iii)~datasets with independent expert-reference AVR measurements, such as INSPIRE-AVR, IOSTAR, AVRDB and Leuven-Haifa, are not currently publicly accessible or require author-mediated access, preventing reference-based calibre validation for either system.

\begin{table}[h]
  \centering
  \caption{Exploratory comparison with VascX on HRF ($N\!=\!30$).}
  \label{tab:vascx}
  \small
\setlength{\tabcolsep}{3pt} 
\renewcommand{\arraystretch}{0.95}
\begin{tabular}{@{}lcccccc@{}}
    \toprule
    Method & $\tau$ & Acc & Sens & Spec & $F_1$ & AUC \\
    \midrule
    VascX \cite{vargas2025vascx} & 0.67 & 0.733 & 1.000 & 0.467 & 0.789 & 0.938 \\
    VascX \cite{vargas2025vascx} & 0.62 & 0.867 & 0.733 & 1.000 & 0.846 & 0.938 \\
    \textbf{Proposed} & \textbf{0.67} & \textbf{0.967} & \textbf{1.000} & \textbf{0.933} & \textbf{0.968} & \textbf{0.978} \\
    \bottomrule
  \end{tabular}
\end{table}

\begin{figure*}[t]
\centering
\includegraphics[width=0.65\textwidth]{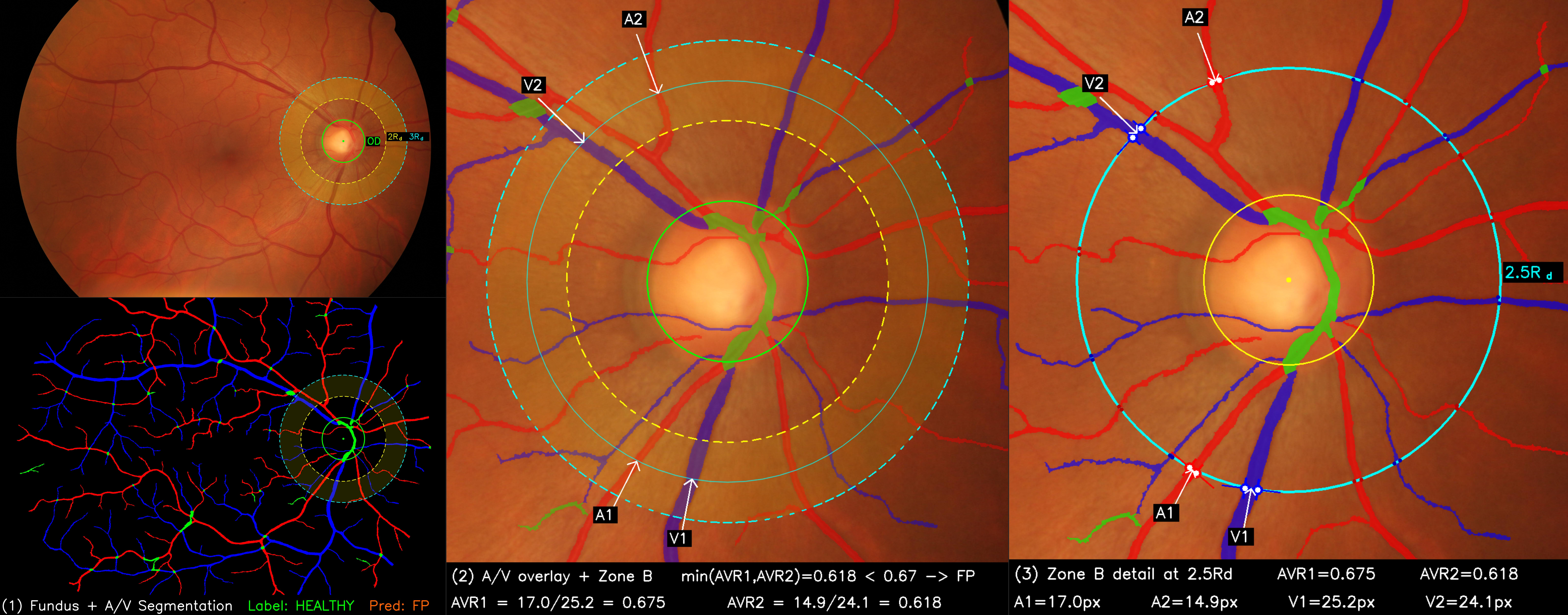}
\caption{Failure case analysis for \texttt{15\_h} (false positive). (1)~Full fundus with OD and Zone~B circles ($2r_d$, $3r_d$) alongside the A/V segmentation. (2)~A/V overlay with arrows marking the two dominant artery and vein crossings (A1, A2, V1, V2). (3)~Zoomed Zone~B calibre measurement at $r=2.5\,r_d$ with caliper-style width annotations. The measured widths (A1\,=\,17.0\,px, A2\,=\,14.9\,px, V1\,=\,25.2\,px, V2\,=\,24.1\,px) yield $\avr_1=0.675$ and $\avr_2=0.618$; since $\min(\avr_1,\avr_2)=0.618<0.67$, the image is flagged as a false positive. The arteries are genuinely narrow; the false positive likely reflects constitutional arteriolar narrowness.}
\label{fig:failure_15h}
\end{figure*}

\subsection{Large-Scale Screening Validation on MESSIDOR} \label{sec:res-messidor} On the full MESSIDOR dataset (1{,}200 fundus images from three
ophthalmology departments), the binary normal-versus-retinopathic task
comprises 546 grade-0 and 254 grade-3 images. At the clinical threshold $\tau = 0.67$, the overall accuracy achieved was
0.948 (95\% interval: $[0.861, 0.982]$) and an F1-score of 0.922
(95\% interval: $[0.813, 0.972]$), suggesting that a procedural
AVR-based pipeline can remain competitive with contemporary end-to-end
deep learning approaches while preserving traceability. These
full-dataset figures are estimates extrapolated from the subset's
per-class error rates via Bayesian resampling rather than direct
measurements, and will be confirmed by evaluation on all 800 images.

\section{Discussion}
\label{sec:discussion}

\subsection{What a transparent pipeline bought us} The primary advantage of our approach is not merely reaching competitive diagnostic accuracy on datasets like MESSIDOR, but gaining knowledge of \emph{how} that diagnosis was reached. Consider an end-to-end classifier that misclassifies an eye. There is no intermediate stage to audit, and saliency maps indicate where the gradient is large rather than which computation failed \cite{rudin2019stop}. Improvement therefore becomes a blind search over architectures and augmentations rather than a targeted fix. By contrast, our pipeline allows a reviewer to trace the 96.7\% accuracy directly back to the physical measurements that enabled it. If the pipeline abstains or misclassifies, a clinician can determine whether the failure arose from optic disc localization, A/V assignment, or calibre measurement.

\subsection{Cross-dataset behaviour}

The pipeline's parameters are geometric quantities defined relative to the
optic disc, so they are invariant to image resolution and, in principle, acquisition device. The learned semantic components inherit whatever domain
sensitivity they have, but the artery--vein model degrades gracefully
($0.960 \rightarrow 0.939$ all-pixel accuracy across the HRF/DRIVE segmentation transfer)
rather than collapsing, and errors surface safely as reduced coverage
and lower vote margins rather than confident wrong diagnoses. We regard
degradation into abstention as the correct failure mode for a screening tool.

\subsection{Failure case analysis}
\label{sec:failure}

To demonstrate the error-attribution capability claimed in Section~I, we examine the pipeline's single misclassification on HRF in detail. Image \texttt{15\_h} is labelled healthy but is predicted retinopathic because both dominant vessel pairings yield $\avr$ below the threshold ($\avr_1\!=\!0.645 < 0.67$, $\avr_2\!=\!0.644 < 0.67$). Because the pipeline is procedural, the source of this false positive can be isolated stage by stage:

\begin{itemize}
  \item \textbf{Stage~2 (A/V segmentation):} The artery--vein mask is anatomically coherent; major vessels are correctly labelled and no artery--vein swap is visible in Zone~B.
  \item \textbf{Stage~3 (Optic disc):} The disc centre and radius are correctly estimated; Zone~B covers the expected peripapillary region.
  \item \textbf{Stage~4 (Calibre):} The measured artery diameters are consistently narrow across 16 of 21 Zone~B radii. The two dominant arterial calibres are ${\sim}30$--$35\%$ smaller than their venular counterparts.
  \item \textbf{Stage~5 (AVR):} The resulting \avr\ values ($\avr_1\!=\!0.645$, $\avr_2\!=\!0.644$, both below~$\tau\!=\!0.67$) faithfully reflect the narrow arteries.
\end{itemize}

The pipeline's output consistently indicates narrow arterioles relative to venules across the majority of Zone~B radii. Whether this reflects genuine constitutional arteriolar narrowness or a systematic measurement artefact cannot be determined from these data alone. This is precisely the kind of ambiguity that a procedural pipeline makes actionable: a clinician reviewing the artefacts can immediately see that the arteries are narrow and decide whether to refer. An end-to-end model producing ``retinopathy, $p\!=\!0.72$'' for this eye would offer no such pathway. Fig.~\ref{fig:failure_15h} shows the stage-wise artefacts for this case. The pixel widths in the caption correspond to a single measurement circle at $r\!=\!2.5\,r_d$; the reported \avr\ values (0.645, 0.644) are median-aggregated across the full Zone~B sweep.

\subsection{Clinical validity and proxy labels}
A potential concern is that \avr\ is primarily associated with
hypertensive retinopathy, whereas the HRF labels denote diabetic
retinopathy. However, a prospective population-based study has shown
that retinal vascular geometry measured from fundus photographs predicts
both the incidence and progression of diabetic retinopathy, beyond
established risk factors such as HbA1c and diabetes
duration~\cite{cheung2017retinal}. Our empirical results support this
link: the pipeline achieves $\text{AUC}=0.978$ on HRF using only
vascular geometry, with the single misclassification (\texttt{15\_h})
attributable to constitutional arteriolar narrowness rather than a
breakdown of the AVR--DR association. Evaluation on a dedicated
hypertensive retinopathy cohort with AVR as primary biomarker is planned as future work.

\section{Limitations} \label{sec:limitations}  Three limitations bound the claims above. \begin{enumerate} \item \textbf{Sample size and proxy labels: }The detailed evaluation cohort on HRF consists of only 30 images, and some broad retinopathy labels may conflate pathologies with differing vascular signatures. Larger clinically graded cohorts are needed for definitive validation. \item  \textbf{Reduced AVR estimator:} Eq.~\eqref{eq:avr} uses the two largest arteries and veins rather than the six-vessel Knudtson recursion \cite{knudtson2003revised}, so the resulting \avr\ is not numerically identical to the quantity reported in normative studies.
\item \textbf{Absence of reference calibre data:} Independent         reference calibre measurements were not available for the datasets used in this evaluation. The pipeline produces plausible \avr\ distributions, but validating individual calibre accuracy and absence of systematic bias will be taken up as future work.\end{enumerate}

\section{Conclusion}
\label{sec:conclusion}

We presented a five-stage procedural pipeline that estimates the
arteriolar-to-venular ratio from a fundus photograph and derives a
retinopathy screening decision from it. The pipeline reconstructs the
clinical grading protocol by combining decoupled vessel segmentation,
optic disc localization, minimal-chord calibre estimation, and clinically
grounded AVR thresholding. Every stage emits a human-readable artefact,
making errors attributable to specific components, and no classification
threshold was artificially tuned on evaluation data. On all 22 LES-AV
images, the pipeline recovered a median \avr\ of 0.818 (95\% CI:
[0.792, 0.978]), consistent with healthy vasculature. On the 30-image
HRF screening cohort, it achieved 96.7\% accuracy with 100\% sensitivity
and 93.3\% specificity, successfully separating healthy and diabetic
retinopathy cohorts. On the MESSIDOR dataset, an estimated accuracy of 0.948 and $F_1\!=\!0.922$ was obtained, which is competitive with other end-to-end deep learning
approaches while preserving full decision traceability. The immediate
next step is to evaluate the pipeline on larger clinical cohorts with
explicit hypertensive retinopathy grading and to explore integrating a
learned optic disc detector for end-to-end procedural automation. By
building screening systems this way, we ensure that every diagnostic
decision is mathematically bound to a verifiable physical quantity.

\bibliographystyle{IEEEtran}
\bibliography{references}

@article{sharan2025avseg,
  title   = {Artery-Vein Segmentation from Fundus Images using Deep Learning},
  author  = {Sharan, S. K. and Sahayam, Subin and Jayaraman, Umarani and Lakshmi Priya, A.},
  journal = {arXiv preprint arXiv:2510.03717},
  year    = {2025},
  url     = {https://arxiv.org/abs/2510.03717}
}

@inproceedings{varshan2023diameter,
  title     = {An Algorithm to Calculate Retinal Vessel Diameter in Fundus Images},
  author    = {Yoga Sri Varshan, V. and Sahayam, Subin and Jayaraman, Umarani},
  booktitle = {Proceedings of the Fourteenth Indian Conference on Computer Vision,
               Graphics and Image Processing (ICVGIP)},
  year      = {2023},
  address   = {Rupnagar, India},
  publisher = {ACM},
  doi       = {10.1145/3627631.3627653}
}

@article{hubbard1999methods,
  title   = {Methods for evaluation of retinal microvascular abnormalities associated
             with hypertension/sclerosis in the Atherosclerosis Risk in Communities Study},
  author  = {Hubbard, Larry D and Brothers, Ronald J and King, William N and
             Clegg, Limin X and Klein, Ronald and Cooper, Larry S and
             Sharrett, A Richey and Davis, Matthew D and Cai, Jianwen},
  journal = {Ophthalmology},
  volume  = {106},
  number  = {12},
  pages   = {2269--2280},
  year    = {1999},
  publisher = {Elsevier}
}

@article{knudtson2003revised,
  title   = {Revised formulas for summarizing retinal vessel diameters},
  author  = {Knudtson, Michael D and Lee, Kristine E and Hubbard, Larry D and
             Wong, Tien Yin and Klein, Ronald and Klein, Barbara E K},
  journal = {Current Eye Research},
  volume  = {27},
  number  = {3},
  pages   = {143--149},
  year    = {2003},
  publisher = {Taylor \& Francis}
}

@article{keith1974some,
  title   = {Some different types of essential hypertension: their course and prognosis},
  author  = {Keith, N M and Wagener, H P and Barker, N W},
  journal = {The American Journal of the Medical Sciences},
  volume  = {268},
  number  = {6},
  pages   = {336--345},
  year    = {1974}
}

@article{wang2025geometry,
  title   = {Retinal vascular geometry and its association to silent brain infarction},
  author  = {Wang, Jing and Chen, Tingli and Wang, Jing and Bian, Xiyun and Zhang, Xiaowei
             and Yang, Xiaolong and Qi, Xing and Li, Yihan and Huang, Haixia
             and Qian, Qian and Yuan, Fei},
  journal = {BMC Ophthalmology},
  volume  = {25},
  pages   = {36},
  year    = {2025},
  doi     = {10.1186/s12886-025-03854-x}
}

@article{cheung2017retinal,
  title   = {Retinal vascular geometry and 6 year incidence and progression
             of diabetic retinopathy},
  author  = {Cheung, Carol Yim-lui and Sabanayagam, Charumathi and Law, Antony Kwan-pui
             and Kumari, Neelam and Ting, Daniel Shu-wei and Tan, Gavin
             and Mitchell, Paul and Cheng, Ching Yu and Wong, Tien Yin},
  journal = {Diabetologia},
  volume  = {60},
  pages   = {1717--1728},
  year    = {2017},
}

@article{rudin2019stop,
  title   = {Stop explaining black box machine learning models for high stakes
             decisions and use interpretable models instead},
  author  = {Rudin, Cynthia},
  journal = {Nature Machine Intelligence},
  volume  = {1},
  number  = {5},
  pages   = {206--215},
  year    = {2019},
  publisher = {Nature Publishing Group}
}

@article{hemelings2019artery,
  title   = {Artery--vein segmentation in fundus images using a fully convolutional network},
  author  = {Hemelings, Ruben and Elen, Bart and Stalmans, Ingeborg and
             Van Keer, Karel and De Boever, Patrick and Blaschko, Matthew B},
  journal = {Computerized Medical Imaging and Graphics},
  volume  = {76},
  pages   = {101636},
  year    = {2019},
  publisher = {Elsevier}
}

@article{geirhos2020shortcut,
  title   = {Shortcut learning in deep neural networks},
  author  = {Geirhos, Robert and Jacobsen, J{\"o}rn-Henrik and Michaelis, Claudio
             and Zemel, Richard and Brendel, Wieland and Bethge, Matthias
             and Wichmann, Felix A},
  journal = {Nature Machine Intelligence},
  volume  = {2},
  number  = {11},
  pages   = {665--673},
  year    = {2020},
  publisher = {Nature Publishing Group}
}

@article{vargas2025vascx,
  title   = {VascX models: Model ensembles for retinal vascular analysis
             from color fundus images},
  author  = {Vargas Quiros, Jose and Liefers, Bart and van Vliet, Karin and
             Klaver, Caroline C W},
  journal = {Translational Vision Science \& Technology},
  volume  = {14},
  number  = {7},
  year    = {2025}
}

@article{staal2004ridge,
  title   = {Ridge-based vessel segmentation in color images of the retina},
  author  = {Staal, Joes and Abr{\`a}moff, Michael D and Niemeijer, Meindert and
             Viergever, Max A and van Ginneken, Bram},
  journal = {IEEE Transactions on Medical Imaging},
  volume  = {23},
  number  = {4},
  pages   = {501--509},
  year    = {2004},
  publisher = {IEEE}
}

@article{budai2013robust,
  title   = {Robust vessel segmentation in fundus images},
  author  = {Budai, Attila and Bock, R{\"u}diger and Maier, Andreas and
             Hornegger, Joachim and Michelson, Georg},
  journal = {International Journal of Biomedical Imaging},
  volume  = {2013},
  pages   = {154860},
  year    = {2013}
}

@inproceedings{orlando2018towards,
  title     = {Towards a glaucoma risk index based on simulated hemodynamics
               from fundus images},
  author    = {Orlando, Jos{\'e} Ignacio and Barbosa Breda, Jo{\~a}o and
               van Keer, Karel and Blaschko, Matthew B and Blanco, Pablo J
               and Bulant, Carlos A},
  booktitle = {Medical Image Computing and Computer-Assisted Intervention (MICCAI)},
  pages     = {65--73},
  year      = {2018},
  publisher = {Springer}
}

@article{decenciere2014feedback,
  title   = {Feedback on a publicly distributed image database: the Messidor database},
  author  = {Decenci{\`e}re, Etienne and Zhang, Xiwei and Cazuguel, Guy and
             La{\"y}, Bruno and Cochener, B{\'e}atrice and Trone, Caroline
             and Gain, Philippe and Ordonez, Richard and Massin, Pascale
             and Erginay, Ali and Charton, B{\'e}atrice and Klein, Jean-Claude},
  journal = {Image Analysis \& Stereology},
  volume  = {33},
  number  = {3},
  pages   = {231--234},
  year    = {2014}
}

\end{document}